\pdfoutput=1
\documentclass[11pt]{article}

\usepackage{ACL2023}

\usepackage{times}
\usepackage{latexsym}

\usepackage[T1]{fontenc}
\usepackage[utf8]{inputenc}

\usepackage{microtype}

\usepackage{inconsolata}

\usepackage{amsfonts,amsmath,amssymb}
\usepackage{balance}
\usepackage{booktabs}
\usepackage{multirow}
\usepackage{graphicx}
\usepackage{enumitem}
\usepackage{subfigure}

\title{Improving Continual Relation Extraction by Distinguishing \\ Analogous Semantics}

\author{
    Wenzheng Zhao$^\dagger$ \quad 
    Yuanning Cui$^\dagger$ \quad 
    Wei Hu$^{\dagger,\,\ddagger,\,}$\thanks{\,\, Corresponding author} \\
    $^\dagger$ State Key Laboratory for Novel Software Technology, Nanjing University, China \\
    $^\ddagger$ National Institute of Healthcare Data Science, Nanjing University, China \\
    \texttt{wzzhao.nju.cs@gmail.com, yncui.nju@gmail.com, whu@nju.edu.cn} 
}

\begin{document}
\maketitle

\begin{abstract}
Continual relation extraction (RE) aims to learn constantly emerging relations while avoiding forgetting the learned relations.
Existing works store a small number of typical samples to re-train the model for alleviating forgetting.
However, repeatedly replaying these samples may cause the overfitting problem.
We conduct an empirical study on existing works and observe that their performance is severely affected by analogous relations.
To address this issue, we propose a novel continual extraction model for analogous relations.
Specifically, we design memory-insensitive relation prototypes and memory augmentation to overcome the overfitting problem.
We also introduce integrated training and focal knowledge distillation to enhance the performance on analogous relations.
Experimental results show the superiority of our model and demonstrate its effectiveness in distinguishing analogous relations and overcoming overfitting.
\end{abstract}

%====================%
\section{Introduction}
\label{sect:introduction}

Relation extraction (RE) aims to detect the relation between two given entities in texts. 
For instance, given a sentence ``\textit{Remixes of tracks from Persona 5 were supervised by Kozuka and original composer Shoji Meguro}'' and an entity pair (\textit{Persona 5}, \textit{Shoji Meguro}), the ``\textit{composer}'' relation is expected to be identified by an RE model. 
Conventional RE task assumes all relations are observed at once, ignoring the fact that new relations continually emerge in the real world. 
To deal with emerging relations, some existing works \cite{wang-etal-2019-sentence, han-etal-2020-continual, CML, cui-etal-2021-refining, zhao-etal-2022-consistent, KIP-Framework, hu-etal-2022-improving, ACA} study continual RE.
In continual RE, new relations and their involved samples continually emerge, and the goal is to classify all observed relations.
Therefore, a continual RE model is expected to be able to learn new relations while retaining the performance on learned relations.

Existing works primarily focus on storing and replaying samples to avoid catastrophic forgetting \cite{cl-survey-overfitting} of the learned relations.
On one hand, considering the limited storage and computational resources, it is impractical to store all training samples and re-train the whole model when new relations emerge. 
On the other hand, replaying a small number of samples every time new relations emerge would make the model prone to overfit the stored samples \cite{Verwimp_2021_ICCV, cl-survey-overfitting}. 
Moreover, existing works simply attribute catastrophic forgetting to the decay of previous knowledge as new relations come but seldom delve deeper into the real causation. 
We conduct an empirical study and find that the severe decay of knowledge among analogous relations is a key factor of catastrophic forgetting.

\begin{table}
\centering
\resizebox{\columnwidth}{!}{
  \begin{tabular}{l|c|crcr}
  \toprule
  \multirow{2}{*}{Models} & \multirow{2}{*}{Max sim.} & \multicolumn{2}{c}{FewRel} & \multicolumn{2}{c}{TACRED} \\
  \cmidrule(lr){3-4} \cmidrule(lr){5-6} & & Accuracy & Drop & Accuracy & Drop \\
  \midrule
  \multirow{3}{*}{CRL} & [0.85, 1.00) & 71.1 & 9.7 & 64.8 & 11.4 \\
                       & [0.70, 0.85) & 78.8 & 5.7 & 76.6 & 5.0 \\
                       & (0.00, 0.70) & 87.9 & 3.2 & 89.6 & 0.6 \\
  \midrule
  \multirow{3}{*}{CRECL} & [0.85, 1.00) & 60.4 & 18.9 & 60.7 & 13.9 \\
                         & [0.70, 0.85) & 78.4 & 6.8 & 70.0 & 8.4 \\
                         & (0.00, 0.70) & 83.0 & 5.1 & 79.9 & 4.3 \\
  \bottomrule
  \end{tabular}}
\caption{Results of our empirical study. We divide all relations into three groups according to their maximum similarity to other relations. ``Accuracy'' indicates the average \textit{accuracy} (\%) of relations after the model finishes learning. ``Drop'' indicates the average \textit{accuracy drop} (\%) from learning the relation for the first time to the learning process finished.}
\label{tab:empirical}
\end{table}

Table \ref{tab:empirical} shows the accuracy and accuracy drop of two existing models on the FewRel \cite{han-etal-2018-fewrel} and TACRED \cite{zhang-etal-2017-position} datasets.
CRL \cite{zhao-etal-2022-consistent} and CRECL \cite{hu-etal-2022-improving} are both state-of-the-art models for continual RE.
All relations in the datasets are divided into three groups according to the maximum cosine similarity of their prototypes to other relation prototypes.
A relation prototype is the overall representation of the relation.
We can observe that the performance on relations with higher similarity is poorer, which is reflected in less accuracy and greater accuracy drop.
Given that a relation pair with high similarity is often analogous to each other, the performance on a relation tends to suffer a significant decline, i.e., catastrophic forgetting, when its analogous relations appear.
For example, the accuracy of the previously learned relation ``location'' drops from 0.98 to 0.6 after learning a new relation ``country of origin''.
Therefore, it is important to maintain knowledge among analogous relations for alleviating catastrophic forgetting.
See Appendix~\ref{appsec:empirical} for more details of our empirical study.

To address the above issues, we propose a novel continual extraction model for analogous relations.
Specifically, we introduce memory-insensitive relation prototypes and memory augmentation to reduce overfitting. 
The memory-insensitive relation prototypes are generated by combining static and dynamic representations, where the static representation is the average of all training samples after first learning a relation, and the dynamic representation is the average of stored samples. 
The memory augmentation replaces entities and concatenates sentences to generate more training samples for replay.
Furthermore, we propose integrated training and focal knowledge distillation to alleviate knowledge forgetting of analogous relations.
The integrated training combines the advantages of two widely-used training methods, which contribute to a more robust feature space and better distinguish analogous relations.
One method uses contrastive learning for training and generates prototypes for relation classification, while the other trains a linear classifier.
The focal knowledge distillation assigns high weights to analogous relations, making the model more focus on maintaining their knowledge. 

Our main contributions are summarized below:
\begin{itemize}  
    \item We explicitly consider the overfitting problem in continual RE, which is often ignored by previous works. 
    We propose memory-insensitive relation prototypes and memory augmentation to alleviate overfitting. 
    
    \item We conduct an empirical study and find that analogous relations are hard to distinguish and their involved knowledge is more easily to be forgotten.
    We propose integrated training and focal knowledge distillation to better distinguish analogous relations.
    
    \item The experimental results on two benchmark datasets demonstrate that our model achieves state-of-the-art accuracy compared with existing works, and better distinguishes analogous relations and overcomes overfitting for continual RE.
    Our source code is available at \url{https://github.com/nju-websoft/CEAR}.
\end{itemize}

%====================%
\section{Related Work}
\label{sect:literature}

Continual learning studies the problem of learning from a continuous stream of data \cite{cl-survey-overfitting}.
The main challenge of continual learning is avoiding catastrophic forgetting of learned knowledge while learning new tasks.
Existing continual learning models can be divided into three categories: regularization-based, dynamic architecture, and memory-based.
The regularization-based models \cite{LwF, EWC} impose constraints on the update of parameters important to previous tasks.
The dynamic architecture models \cite{PackNet, BNS} dynamically extend the model architecture to learn new tasks and prevent forgetting previous tasks.
The memory-based models \cite{GEM, iCaRL, A-GEM} store a limited subset of samples in previous tasks and replay them when learning new tasks.

\begin{figure*}[ht]
    \centering
    \includegraphics[width=\linewidth]{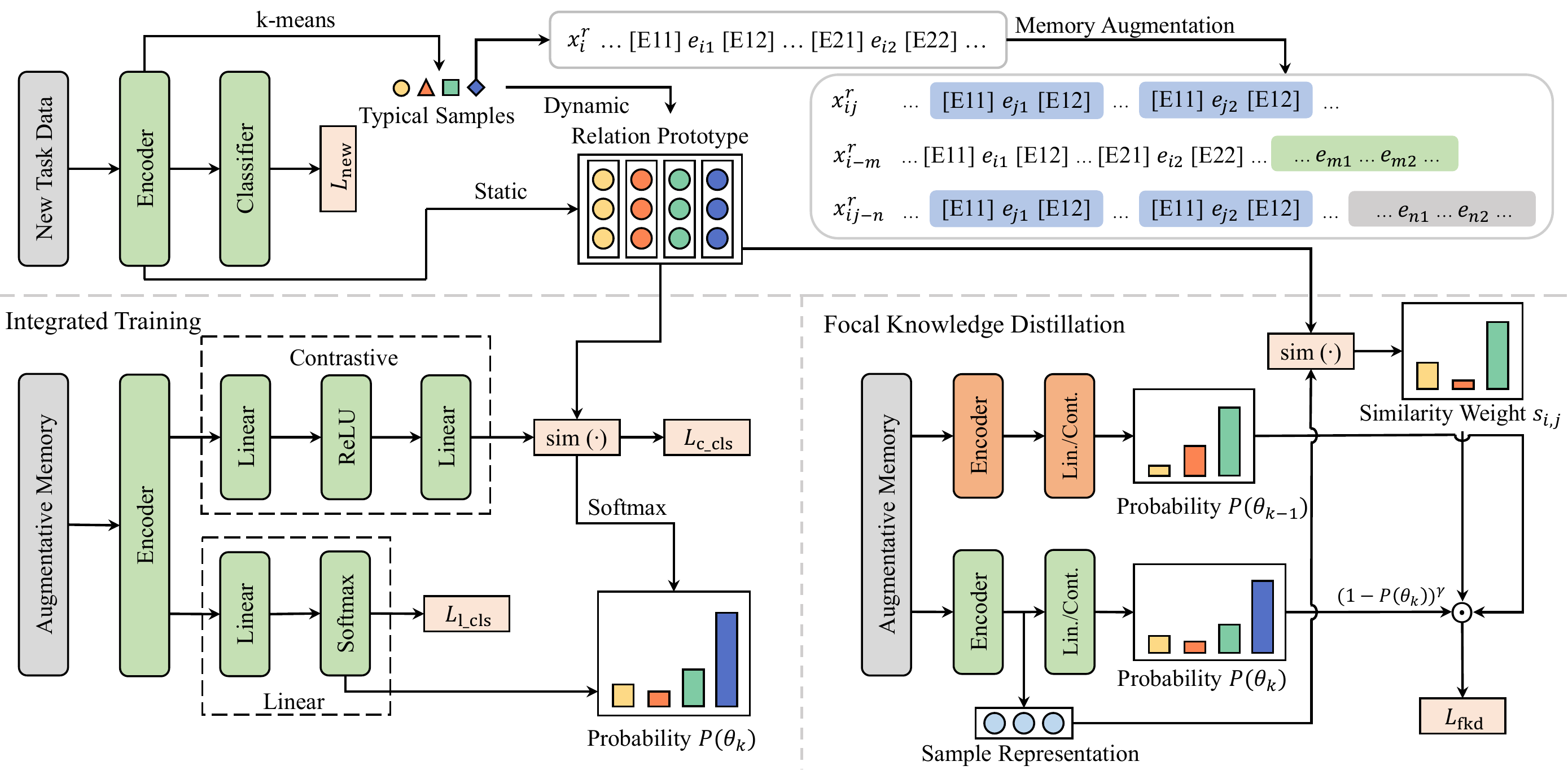}
    \caption{Framework of the proposed model for task $T_k$.}
    \label{framework}
\end{figure*}

In continual RE, the memory-based models \cite{wang-etal-2019-sentence, han-etal-2020-continual, CML, cui-etal-2021-refining, zhao-etal-2022-consistent, KIP-Framework, hu-etal-2022-improving} are the mainstream choice as they have shown better performance for continual RE than others.
To alleviate catastrophic forgetting, previous works make full use of relation prototypes, contrastive learning, multi-head attention, knowledge distillation, etc.
EA-EMR \cite{wang-etal-2019-sentence} introduces memory replay and the embedding aligned mechanism to mitigate the embedding distortion when training new tasks.
CML \cite{CML} combines curriculum learning and meta-learning to tackle the order sensitivity in continual RE.
RP-CRE \cite{cui-etal-2021-refining} and KIP-Framework \cite{KIP-Framework} leverage relation prototypes to refine sample representations through multi-head attention-based memory networks.
Additionally, KIP-Framework uses external knowledge to enhance the model through a knowledge-infused prompt to guide relation prototype generation.
EMAR \cite{han-etal-2020-continual}, CRL \cite{zhao-etal-2022-consistent}, and CRECL \cite{hu-etal-2022-improving} leverage contrastive learning for model training.
Besides, knowledge distillation is employed by CRL to maintain previously learned knowledge.
ACA \cite{ACA} is the only work that considers the knowledge forgetting of analogous relations ignored by the above works and proposes an adversarial class augmentation strategy to enhance other continual RE models.
All these models do not explicitly consider the overfitting problem \cite{cl-survey-overfitting, Verwimp_2021_ICCV}, which widely exists in the memory-based models.
As far as we know, a few works \cite{wang2021revisiting} in other continual learning fields have tried to reduce the overfitting problem and achieve good results.
We address both the problems of distinguishing analogous relations and overfitting to stored samples, and propose an end-to-end model.

%====================%
\section{Task Definition}
\label{sect:definition}

A continual RE task consists of a sequence of tasks $\mathcal{T}=\{T_1, T_2, \dots, T_K\}$. 
Each individual task is a conventional RE task.
Given a sentence, the RE task aims to find the relation between two entities in this sentence. 
The dataset and relation set of $T_k\in\mathcal{T}$ are denoted by $D_k$ and $R_k$, respectively.
$D_k$ contains separated training, validation and test sets, denoted by $D_k^{\rm train}$, $D_k^{\rm valid}$ and $D_k^{\rm test}$, respectively. 
$R_k$ contains at least one relation. 
The relation sets of different tasks are disjoint. 

Continual RE aims to train a classification model that performs well on both current task $T_k$ and previously accumulated tasks $\tilde{T}_{k-1} = \bigcup_{i=1}^{k-1} T_i$. 
In other words, a continual RE model is expected to be capable of identifying all seen relations $\tilde{R}_k=\bigcup_{i=1}^k R_i$ and would be evaluated on all the test sets of seen tasks $\tilde{D}_k^{\rm test}=\bigcup_{i=1}^k D_i^{\rm test}$.

%====================%
\section{Methodology}
\label{sect:methodology}

%--------------------%
\subsection{Overall Framework}

The overall framework is shown in Figure \ref{framework}. 
For a new task $T_k$, we first train the continual RE model on $D_k$ to learn this new task. 
Then, we select and store a few typical samples for each relation $r\in R_k$.
Next, we calculate the prototype $\mathbf{p}_r$ of each relation $r\in\tilde{R}_k$ according to the static and dynamic representations of samples.
We also conduct memory augmentation to provide more training data for memory replay.
Note that the augmented data are not used for prototype generation.
Finally, we perform memory replay consisting of integrated training and focal knowledge distillation to alleviate catastrophic forgetting.
The parameters are updated in the first and last steps.
After learning $T_k$, the model continually learns the next task $T_{k+1}$.

%--------------------%
\subsection{New Task Training}

When the new task $T_k$ emerges, we first train the model on $D_k^\text{train}$.
We follow the works \cite{cui-etal-2021-refining, zhao-etal-2022-consistent, KIP-Framework, hu-etal-2022-improving} to use the pre-trained language model BERT \cite{devlin-etal-2019-bert} as the encoder.

Given a sentence $x$ as input, we first tokenize it and insert special tokens $[E_{11}]/[E_{12}]$ and $[E_{21}]/[E_{22}]$ to mark the start/end positions of head and tail entities, respectively.
We use the hidden representations of $[E_{11}]$ and $[E_{21}]$ as the representations of head and tail entities.
The representation of $x$ is defined as
\begin{align}
\mathbf{h}_x = \mathrm{LayerNorm}\big(\mathbf{W}_1 [\mathbf{h}^{11}_x;\mathbf{h}_x^{21}] + \mathbf{b}\big),
\end{align}
where $\mathbf{h}_x^{11}, \mathbf{h}_x^{21} \in \mathbb{R}^d $ are the hidden representations of head and tail entities, respectively. 
$d$ is the dimension of the hidden layer in BERT.
$\mathbf{W}_1 \in \mathbb{R}^{d \times 2d}$ and $\mathbf{b} \in \mathbb{R}^d$ are two trainable parameters.

Then, we use a linear softmax classifier to calculate the classification probability of $x$ according to the representation $\mathbf{h}_x$:
\begin{align}
    P(x; \theta_k) = \mathrm{softmax}(\mathbf{W}_2 \mathbf{h}_x),
\end{align}
where $\theta_k$ denotes the model when learning $T_k$.
$\mathbf{W}_2 \in \mathbb{R}^{|\tilde{R}_k| \times d}$ is the trainable parameter of the linear classifier.

Finally, the classification loss of new task training is calculated as follows:
\begin{align}
\resizebox{\columnwidth}{!}{$
\mathcal{L}_\text{new} = -\frac{1}{|D_k^\text{train}|} \sum\limits_{x_i \in D_k^\text{train}} \sum\limits_{r_j \in R_k} \delta_{y_i, r_j} \log P(r_j\,|\,x_i; \theta_k),
$}\end{align}
where $P(r_j\,|\,x_i; \theta_k)$ is the probability of input $x_i$ classified as relation $r_j$ by the current model $\theta_k$. 
$y_i$ is the label of $x_i$ such that if $y_i=r_j, \delta_{y_i, r_j} = 1$, and 0 otherwise.

%--------------------%
\subsection{Memory Sample Selection}

To preserve the learned knowledge from previous tasks, we select and store a few typical samples for memory replay.
Inspired by the works \cite{han-etal-2020-continual, cui-etal-2021-refining, zhao-etal-2022-consistent, KIP-Framework, hu-etal-2022-improving}, we adopt the k-means algorithm to cluster the samples of each relation $r\in R_k$. 
The number of clusters is defined as the memory size $m$. 
For each cluster, we select the sample whose representation is closest to the medoid and store it in the memory space $M^r$.
The accumulated memory space is $\tilde{M}_k = \bigcup_{r \in \tilde{R}_k} M^r$.

%--------------------%
\subsection{Memory-Insensitive Relation Prototype} 

A relation prototype is the overall representation of the relation.
Several previous works \cite{han-etal-2020-continual, zhao-etal-2022-consistent, hu-etal-2022-improving} directly use relation prototypes for classification and simply calculate the prototype of $r$ using the average of the representations of its typical samples.
But, such a relation prototype is sensitive to the typical samples, which may cause the overfitting problem.
To reduce the sensitivity to typical samples, \citet{KIP-Framework} propose a knowledge-infused relation prototype generation, which employs a knowledge-infused prompt to guide prototype generation. 
However, it relies on external knowledge and thus brings additional computation overhead.

To alleviate the overfitting problem, we first calculate and store the average representation of all training samples after first learning a relation.
This representation contains more comprehensive knowledge about the relation.
However, as we cannot store all training samples, it is \emph{static} and cannot be updated to adapt to the new feature space in the subsequent learning.
In this paper, the \emph{dynamic} representation of typical samples is used to finetune the \emph{static} representation for adapting the new feature space.
The memory-insensitive relation prototype of relation $r$ is calculated as follows:
\begin{align}
\mathbf{p}_r = (1-\beta)\,\mathbf{p}_r^\text{static} + \frac{\beta}{|M^r|} \sum_{x_i \in M^r} \mathbf{h}_{x_i},
\end{align}
where $\mathbf{p}_r^\text{static}$ is the average representation of all training samples after learning relation $r$ for the first time, and $\beta$ is a hyperparameter.

%--------------------%
\subsection{Memory Augmentation}

The memory-based models \cite{wang-etal-2019-sentence, han-etal-2020-continual, cui-etal-2021-refining, zhao-etal-2022-consistent, KIP-Framework, hu-etal-2022-improving} select and store a small number of typical samples and replay them in the subsequent learning. 
Due to the limited memory space, these samples may be replayed many times during continual learning, resulting in overfitting.
To address this issue, we propose a memory augmentation strategy to provide more training samples for memory replay.

For a sample $x^r_i$ of relation $r$ in $M^r$, we randomly select another sample $x^r_j \neq x^r_i$ from $M^r$. 
Then, the head and tail entities of $x^r_i$ are replaced by the corresponding entities of $x^r_j$ and the new sample, denoted by $x^r_{ij}$, can be seen as an additional sample of relation $r$.
Also, we use sentence concatenation to generate training samples.
Specifically, we randomly select another two samples $x_m$ and $x_n$ from $\tilde{M}_k \setminus M^r$ and append them to the end of $x^r_i$ and $x^r_{ij}$, respectively.
Note that $x_m$ and $x_n$ are not the typical samples of relation $r$.
Then, we obtain two new samples of relation $r$, denoted by $x^r_{i-m}$ and $x^r_{ij-n}$.
The model is expected to still identify the relation $r$ though there is an irrelevant sentence contained in the whole input.
We conduct this augmentation strategy on all typical samples in $\tilde{M}_k$, but the augmented data are only used for training, not for prototype generation, as they are not accurate enough.
Finally, the overall augmented memory space is $\hat{M}_k$, and $|\hat{M}_k| = 4|\tilde{M}_k|$.

%--------------------%
\subsection{Memory Replay}

\subsubsection{Integrated Training}

There are two widely-used training methods for continual RE: \citet{han-etal-2020-continual, zhao-etal-2022-consistent, hu-etal-2022-improving} use contrastive learning for training and make predictions via relation prototypes; \citet{cui-etal-2021-refining, KIP-Framework} leverage the cross entropy loss to train the encoder and linear classifier.
We call these two methods the \emph{contrastive} method and the \emph{linear} method, respectively.

The contrastive method contributes to a better feature space because it pulls the representations of samples from the same relation and pushes away those from different relations, which improves the alignment and uniformity \cite{pmlr-v119-wang20k}.
However, its prediction process is sensitive to the relation prototypes, especially those of analogous relations that are highly similar to each other.
The linear classifier decouples the representation and classification processes, which ensures a more task-specific decision boundary.
We adopt both contrastive and linear methods to combine their merits:
\begin{align}
\mathcal{L}_\text{cls} = \mathcal{L}_\text{c\_cls} + \mathcal{L}_\text{l\_cls},
\end{align}
where $\mathcal{L}_\text{c\_cls}$ and $\mathcal{L}_\text{l\_cls}$ denote the losses of the contrastive and linear methods, respectively.

In the contrastive method, we first leverage two-layer MLP to reduce dimension:
\begin{align}
    \mathbf{z}_x = \mathrm{Norm} \big( \mathrm{MLP}(\mathbf{h}_x) \big).
\end{align}

Then, we use the InfoNCE loss \cite{infonce-loss} and the triplet loss \cite{triplet-loss} in contrastive learning:
\begin{align}
\resizebox{\columnwidth}{!}{$ 
\begin{aligned}
\mathcal{L}_\text{c\_cls} 
     =& - \frac{1}{|\hat{M}_k|} \sum_{x_i\in\hat{M}_k} \log \frac{\exp(\mathbf{z}_{x_i} \cdot \mathbf{z}_{y_i} / \tau_1)}{\sum_{r \in \tilde{R}_k} \exp(\mathbf{z}_{x_i} \cdot \mathbf{z}_r / \tau_1)} \\
& + \frac{\mu}{|\hat{M}_k|} \sum_{x_i\in\hat{M}_k} \max(\omega - \mathbf{z}_{x_i} \mathbf{z}_{y_i} + \mathbf{z}_{x_i} \mathbf{z}_{y^{\prime}_i}, 0)
\end{aligned},$}
\end{align}
where $\mathbf{z}_r$ is the low-dimensional prototype of relation $r$.
$y^{\prime}_i = \mathop{\arg\max}_{y^{\prime}_i \in \tilde{R}_k \setminus \{y_i\}} \mathbf{z}_{x_i} \cdot \mathbf{z}_{y^{\prime}_i}$ is the most similar negative relation label of sample $x_i$. 
$\tau_1$ is the temperature parameter.
$\mu$ and $\omega$ are hyperparameters.

At last, the relation probability is computed through the similarity between the representations of test sample and relation prototypes:
\begin{align}
    P_c(x_i; \theta_k) = \mathrm{softmax} (\mathbf{z}_{x_i} \cdot \mathbf{Z}_{\tilde{R}_k}),
\end{align}
where $\mathbf{Z}_{\tilde{R}_k}$ denotes the matrix of prototypes of all seen relations.

In the linear method, a linear classifier obtains the relation probability similar to that in the new task training step. 
The loss function is
\begin{align}
\resizebox{\columnwidth}{!}{$
\mathcal{L}_\text{l\_cls} = -\frac{1}{|\hat{M}_k|} \sum\limits_{x_i\in\hat{M}_k} \sum\limits_{r_j\in\tilde{R}_k} \delta_{y_i, r_j} \log P(r_j\,|\,x_i; \theta_k).
$}\end{align}

\subsubsection{Focal Knowledge Distillation}

During the continual training process, some emerging relations are similar to other learned relations and are difficult to distinguish.
Inspired by the focal loss \cite{focal-loss}, we propose the focal knowledge distillation, which forces the model to focus more on analogous relations.

Specifically, we assign a unique weight for each sample-relation pair, according to the classification probability of the sample and the similarity between the representations of sample and relation prototype. 
Difficult samples and analogous sample-relation pairs are assigned high weights.
The weight $w_{i, j}$ for sample $x_i$ and relation $r_j$ is
\begin{align}
    s_{x_i, r_j} &= \frac{\exp \big(\mathrm{sim}(\mathbf{h}_{x_i}, \mathbf{p}_{r_j}) / \tau_2\big)}{\sum_{ r_m \in \tilde{R}_{k-1}} \exp \big(\mathrm{sim}(\mathbf{h}_{x_i}, \mathbf{p}_{r_m}) / \tau_2\big)}, \\
    w_{x_i, r_j} &= s_{x_i, r_j} \big(1-P(y_i\,|\,x_i; \theta_k)\big)^\gamma,
\end{align}
where $\mathbf{p}_{r_j}$ is the prototype of relation $r_j$. 
$\mathrm{sim}(\cdot)$ is the similarity function, e.g., cosine.
$\tau_2$ is the temperature parameter and $\gamma$ is a hyperparameter.

With $w_{x_i, r_j}$, the focal knowledge distillation loss is calculated as follows:
\begin{align}
&a_{x_i, r_j} = w_{x_i, r_j} P(r_j\,|\,x_i; \theta_{k-1}),\\
&\resizebox{\columnwidth}{!}{$
\mathcal{L}_\text{fkd} = - \frac{1}{|\hat{M}_k|} \sum\limits_{x_i\in\hat{M}_k} \sum\limits_{r_j\in\tilde{R}_{k-1}} a_{x_i, r_j} \log P(r_j\,|\,x_i; \theta_k),
$}\end{align}
where $P(r_j\,|\,x_i; \theta_{k-1})$ denotes the probability of sample $x_i$ predicted to relation $r_j$ by the previous model $\theta_{k-1}$.

The focal knowledge distillation loss is combined with the training losses of contrastive and linear methods. 
The overall loss is defined as
\begin{align}
    \mathcal{L}_\text{replay} = \mathcal{L}_\text{cls} + \lambda_{1} \mathcal{L}_\text{c\_fkd} + \lambda_2 \mathcal{L}_\text{l\_fkd},
\end{align}
where $\mathcal{L}_\text{c\_fkd}$ and $\mathcal{L}_\text{l\_fkd}$ are the focal knowledge distillation losses of contrastive and linear methods, respectively. 
$\lambda_1$ and $\lambda_2$ are hyperparameters.

\begin{table*}
\centering
\resizebox{\textwidth}{!}{
\begin{tabular}{l|cccccccccc}
\toprule
  \textbf{FewRel} & $T_1$ & $T_2$ & $T_3$ & $T_4$ & $T_5$ & $T_6$ & $T_7$ & $T_8$ & $T_9$ & $T_{10}$ \\ 
\midrule
  EA-EMR & 89.0 & 69.0 & 59.1 & 54.2 & 47.8 & 46.1 & 43.1 & 40.7 & 38.6 & 35.2 \\
  EMAR (BERT) & \textbf{98.8} & 89.1 & 89.5 & 85.7 & 83.6 & 84.8 & 79.3 & 80.0 & 77.1 & 73.8 \\
  CML & 91.2 & 74.8 & 68.2 & 58.2 & 53.7 & 50.4 & 47.8 & 44.4 & 43.1 & 39.7 \\
  RP-CRE & 97.9 & 92.7 & 91.6 & 89.2 & 88.4 & 86.8 & 85.1 & 84.1 & 82.2 & 81.5 \\ 
  CRL & 98.2 & 94.6 & 92.5 & 90.5 & 89.4 & 87.9 & \underline{86.9} & \underline{85.6} & \underline{84.5} & \underline{83.1} \\
  CRECL & 97.8 & \underline{94.9} & \underline{92.7} & 90.9 & 89.4 & 87.5 & 85.7 & 84.6 & 83.6 & 82.7 \\
  $\text{KIP-Framework}^\triangle$ & \underline{98.4} & 93.5 & 92.0 & \underline{91.2} & \underline{90.0} & \underline{88.2} & \underline{86.9} & \underline{85.6} & 84.1 & 82.5 \\ 
\midrule
  Ours & 98.1$_{\,\pm 0.6}$ & \textbf{95.8}$_{\,\pm 1.7}$ & \textbf{93.6}$_{\,\pm 2.1}$ & \textbf{91.9}$_{\,\pm 2.0}$ & \textbf{91.1}$_{\,\pm 1.5}$ & \textbf{89.4}$_{\,\pm 2.0}$ & \textbf{88.1}$_{\,\pm 0.7}$ & \textbf{86.9}$_{\,\pm 1.3}$ & \textbf{85.6}$_{\,\pm 0.8}$ & \textbf{84.2}$_{\,\pm 0.4}$ \\
\bottomrule
\toprule
  \textbf{TACRED} & $T_1$ & $T_2$ & $T_3$ & $T_4$ & $T_5$ & $T_6$ & $T_7$ & $T_8$ & $T_9$ & $T_{10}$ \\ 
\midrule
  EA-EMR & 47.5 & 40.1 & 38.3 & 29.9 & 28.4 & 27.3 & 26.9 & 25.8 & 22.9 & 19.8 \\
  EMAR (BERT) & 96.6 & 85.7 & 81.0 & 78.6 & 73.9 & 72.3 & 71.7 & 72.2 & 72.6 & 71.0 \\
  CML & 57.2 & 51.4 & 41.3 & 39.3 & 35.9 & 28.9 & 27.3 & 26.9 & 24.8 & 23.4 \\
  RP-CRE & 97.6 & 90.6 & 86.1 & 82.4 & 79.8 & 77.2 & 75.1 & 73.7 & 72.4 & 72.4 \\
  CRL & \underline{97.7} & 93.2 & 89.8 & 84.7 & 84.1 & 81.3 & 80.2 & 79.1 & 79.0 & 78.0 \\
  CRECL & 96.6 & 93.1 & 89.7 & \underline{87.8} & \underline{85.6} & \underline{84.3} & \textbf{83.6} & \textbf{81.4} & 79.3 & 78.5 \\ 
  $\text{KIP-Framework}^\triangle$ & \textbf{98.3} & \textbf{95.0} & \underline{90.8} & 87.5 & 85.3 & \underline{84.3} & 82.1 & 80.2 & \underline{79.6} & \underline{78.6} \\ 
\midrule
  Ours & \underline{97.7}$_{\,\pm 1.6}$ & \underline{94.3}$_{\,\pm 2.9}$ & \textbf{92.3}$_{\,\pm 3.3}$ & \textbf{88.4}$_{\,\pm 3.7}$ & \textbf{86.6}$_{\,\pm 3.0}$ & \textbf{84.5}$_{\,\pm 2.1}$ & \underline{82.2}$_{\,\pm 2.8}$ & \underline{81.1}$_{\,\pm 1.6}$ & \textbf{80.1}$_{\,\pm 0.7}$ & \textbf{79.1}$_{\,\pm 1.1}$ \\ 
\bottomrule
\end{tabular}}
\caption{Accuracy (\%) on all observed relations after learning each task. 
The best results are marked in bold, and the second-best ones are marked with underlines. 
``$\triangle$'' indicates the model using external knowledge.}
\label{tab:main-result}
\end{table*}

%--------------------%
\subsection{Relation Prediction}

After learning task $T_k$, the contrastive and linear methods are combined to predict the relation label of the given test sample $x_i^*$:
\begin{align}
\resizebox{\columnwidth}{!}{$ 
    y_i^* = \mathop{\arg\max}\limits_{y_i^* \in \tilde{R}_k}
    \big((1-\alpha) P_c(x_i^*; \theta_k) + \alpha P_l(x_i^*; \theta_k)\big),
$}\end{align}
where $P_c(x_i^*; \theta_k)$ and $P_l(x_i^*; \theta_k)$ are the probabilities calculated by the contrastive and linear methods, respectively.
$\alpha$ is a hyperparameter.

%====================%
\section{Experiments and Results}
\label{sect:exp}

In this section, we report the experimental results of our model. 
The source code is accessible online.

%--------------------%
\subsection{Datasets}

We conduct our experiments on two widely-used benchmark datasets:
\begin{itemize}
\item\textbf{FewRel} \cite{han-etal-2018-fewrel} is a popular RE dataset originally built for few-shot learning. 
It contains 100 relations and 70,000 samples in total. 
To be in accord with previous works \cite{cui-etal-2021-refining,zhao-etal-2022-consistent}, we use 80 relations each with 700 samples (i.e., in the training and validation sets), and split them into 10 subsets to simulate 10 disjoint tasks.

\item\textbf{TACRED} \cite{zhang-etal-2017-position} is a large-scale RE dataset having 42 relations and 106,264 samples. 
Following the experiment setting of previous works, we remove ``\textit{no\_relation}'' and divide other relations into 10 tasks. 
\end{itemize}

%--------------------%
\subsection{Experiment Setting and Baseline Models}

RP-CRE \cite{cui-etal-2021-refining} proposes a completely-random strategy to split all relations into 10 subsets corresponding to 10 tasks, and \emph{accuracy} on all observed relations is chosen as the evaluation metric, which is defined as the proportion of correctly predicted samples in the whole test set.
This setting is widely followed by existing works \cite{zhao-etal-2022-consistent, KIP-Framework, hu-etal-2022-improving}. 
For a fair comparison, we employ the same setting and obtain the divided data from the open-source code of RP-CRE to guarantee exactly the same task sequence. 
Again, following existing works, we carry out the main experiment with a memory size of 10 and report the average result of five different task sequences. 
See Appendix \ref{appsec:hyperparameters} for the details of the hyperparameter setting.

For comparison, we consider the following baseline models: EA-EMR \cite{wang-etal-2019-sentence}, EMAR \cite{han-etal-2020-continual}, CML \cite{CML}, RP-CRE \cite{cui-etal-2021-refining}, CRL \cite{zhao-etal-2022-consistent}, CRECL \cite{hu-etal-2022-improving} and KIP-Framework \cite{KIP-Framework}.
See Section~\ref{sect:literature} for their details.

%--------------------%
\subsection{Results and Analyses}

\subsubsection{Main Results}

Table \ref{tab:main-result} shows the results of all compared baselines in the main experiment.
% All of the reported results are the average of the same 5 task sequences.
The results of EA-EMR, EMAR, CML, and RP-CRE are obtained from the RP-CRE's original paper, and the results of other baselines are directly cited from their original papers.
We additionally report the standard deviations of our model.
Based on the results, the following observations can be drawn:

Our proposed model achieves an overall state-of-the-art performance on the two different datasets for the reason that our model can reduce overfitting to typical samples and better maintain knowledge among analogous relations.
Thus, we can conclude that our model effectively alleviates catastrophic forgetting in continual RE.

As new tasks continually emerge, the performance of all compared models declines, which indicates that catastrophic forgetting is still a major challenge to continual RE.
EA-EMR and CML do not use BERT as the encoder, so they suffer the most performance decay.
This demonstrates that BERT has strong stability for continual RE.

All models perform relatively poorer on TACRED and the standard deviations of our model on TACRED are also higher than those on FewRel.
The primary reason is that TACRED is class-imbalanced and contains fewer training samples for each relation.
Therefore, it is more difficult and leads to greater randomness in the task division.

\subsubsection{Ablation Study}

We conduct an ablation study to validate the effectiveness of individual modules in our model.
Specifically, for ``w/o FKD'', we remove the focal knowledge distillation loss in memory replay; 
for ``w/o LM'' or ``w/o CM'', the model is only trained and evaluated with the contrastive or linear method; 
for ``w/o MA'', we only train the model with original typical samples in memory replay; 
and for ``w/o DP'' or ``w/o SP'', we directly generate relation prototypes based on the average of static or dynamic representations.

The results are shown in Table \ref{tab:ablation-study}.
It is observed that our model has a performance decline without each component, which demonstrates that all modules are necessary.
Furthermore, the proposed modules obtain greater improvement on the TACRED dataset.
The reason is that TACRED is more difficult than FewRel, so the proposed modules are more effective in difficult cases.

\subsubsection{Influence of Memory Size}

\begin{figure*}
    \centering
    \includegraphics[width=\linewidth]{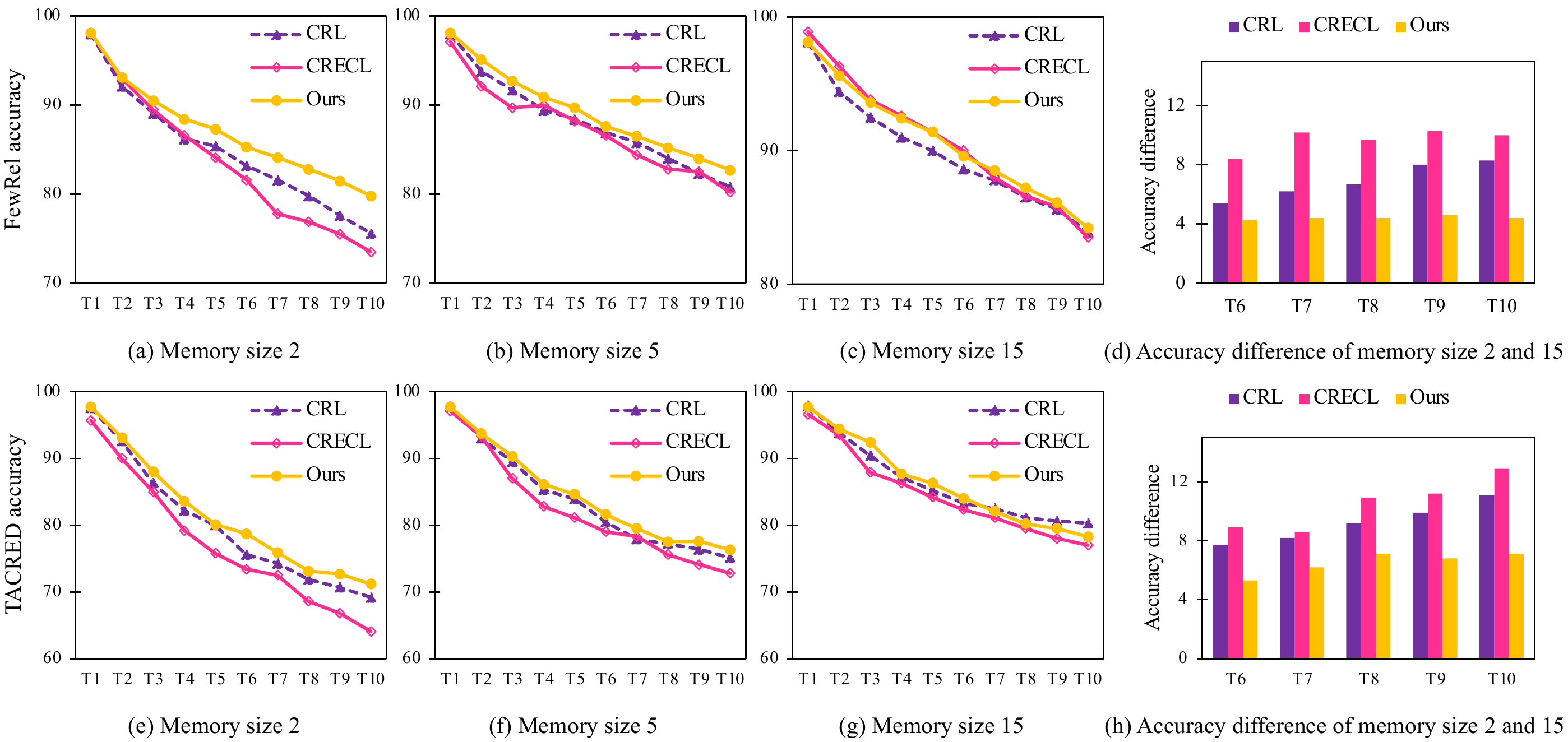}
    \caption{Accuracy w.r.t. different memory sizes and accuracy difference between memory sizes.}
    \label{fig:memsize}
\end{figure*}

\begin{table}[!tb]
{\small
\begin{tabular}{c|l|ccccc}
\toprule
  \multicolumn{2}{c|}{} & $T_6$ & $T_7$ & $T_8$ & $T_9$ & $T_{10}$ \\ 
\midrule
  \parbox[t]{3mm}{\multirow{6}{*}{\rotatebox[origin=c]{90}{FewRel}}} 
  & Intact Model & \textbf{89.4} & \textbf{88.1} & \textbf{86.9} & \textbf{85.6} & \textbf{84.2} \\
  & \ \ w/o FKD   & 89.3 & 88.0 & 86.8 & 85.5 & 84.0 \\
  & \ \ w/o LM    & 89.0 & 87.5 & 86.5 & 85.1 & 83.6 \\
  & \ \ w/o CM    & 89.3 & 87.5 & 86.8 & \textbf{85.6} & 84.0 \\
  & \ \ w/o MA    & 88.4 & 87.4 & 86.4 & 85.4 & 83.7 \\
  & \ \ w/o DP    & 89.2 & 87.9 & 86.6 & 85.3 & 83.8 \\
  & \ \ w/o SP    & 89.3 & 87.8 & 86.6 & 85.2 & 83.5 \\
\midrule
  \parbox[t]{3mm}{\multirow{6}{*}{\rotatebox[origin=c]{90}{TACRED}}} 
  & Intact Model & \textbf{84.5} & \textbf{82.2} & \textbf{81.1} & \textbf{80.1} & \textbf{79.1} \\
  & \ \ w/o FKD   & 83.4 & 81.3 & 79.5 & 79.2 & 78.2 \\
  & \ \ w/o LM    & 83.7 & 81.2 & 79.6 & 79.4 & 78.2 \\
  & \ \ w/o CM    & 84.0 & 81.9 & 80.1 & 79.2 & 78.0 \\
  & \ \ w/o MA    & 82.9 & 81.2 & 79.3 & 79.0 & 77.9 \\
  & \ \ w/o DP    & 83.2 & 80.8 & 79.1 & 79.1 & 78.3 \\
  & \ \ w/o SP    & 83.5 & 81.1 & 79.6 & 79.3 & 78.2 \\
\bottomrule
\end{tabular}}
\caption{Ablation study results. We remove focal knowledge distillation (FKD), linear method (LM), contrastive method (CM), memory augmentation (MA), dynamic prototypes (DP), and static prototypes (SP) in order and report the accuracy (\%) on all observed relations.}
\label{tab:ablation-study}
\end{table}

Memory size is defined as the number of stored typical samples for each relation.
For the memory-based models in continual RE, their performance is highly influenced by memory size.
We conduct an experiment with different memory sizes to compare our model with CRL and CRECL for demonstrating that our model is less sensitive to memory size.
We re-run the source code of CRL and CRECL with different memory sizes and show the results in Figure~\ref{fig:memsize}.
Note that we do not compare with KIP-Framework because it uses external knowledge to enhance performance, which is beyond our scope.

In most cases, our model achieves state-of-the-art performance with different memory sizes, which demonstrates the strong generalization of our model.
However, our model does not obtain the best performance on TACRED with memory size 15 because the overfitting problem that we consider is not serious in this case.
In fact, as the memory size becomes smaller, the overfitting problem is getting worse, and analogous relations are more difficult to distinguish due to the limited training data samples.
From Figures~\ref{fig:memsize}(a), (b), (e), and (f), our model has greater advantages when the memory size is small, which indicates that our model can better deal with the overfitting problem in continual RE.

We also observe that the performance of each model declines due to the decrease of memory size, which demonstrates that memory size is a key factor in the performance of continual RE models.
From Figures~\ref{fig:memsize}(d) and (h), the performance difference between different memory sizes is smaller.
Thus, we draw the conclusion that our model is more robust to the change of memory size.

\subsubsection{Performance on Analogous Relations}

One strength of our model is to distinguish analogous relations for continual RE.
We conduct an experiment to explore this point.
Specifically, we select relations in the former five tasks which have analogous ones in the latter tasks, and report the accuracy and drop on them in Table \ref{tab:ana-relations}.
We consider that two relations are analogous if the similarity between their prototypes is greater than 0.85.
As aforementioned, knowledge of the relations is more likely to be forgotten when their analogous relations emerge.
Thus, all compared models are challenged by these relations.
However, the performance of our model is superior and drops the least, which shows that our model succeeds in alleviating knowledge forgetting among analogous relations.

\begin{table}[!tb]
\centering
{\small
\begin{tabular}{l|cc|cc}
\toprule
\multirow{2}{*}{Models} & \multicolumn{2}{c|}{FewRel} & \multicolumn{2}{c}{TACRED} \\
\cmidrule(lr){2-3} \cmidrule(lr){4-5} & Accuracy & Drop & Accuracy & Drop \\
\midrule
CRL   & 69.7 & 19.0 & 68.9 & 20.4 \\
CRECL & 66.0 & 23.6 & 62.3 & 25.3 \\
Ours  & \textbf{71.1} & \textbf{18.7} & \textbf{70.4} & \textbf{18.3} \\
\bottomrule
\end{tabular}}
\caption{Accuracy (\%) and accuracy drop (\%) on analogous relations. 
We select relations in the former five tasks that have similar ones in the latter tasks. 
Accuracy and drop are calculated in the same way as Table \ref{tab:empirical}.}
\label{tab:ana-relations}
\end{table}

\subsubsection{Case Study}
\label{sec:case-study}

\begin{figure}[!tb]
    \centering
    \includegraphics[width=\linewidth]{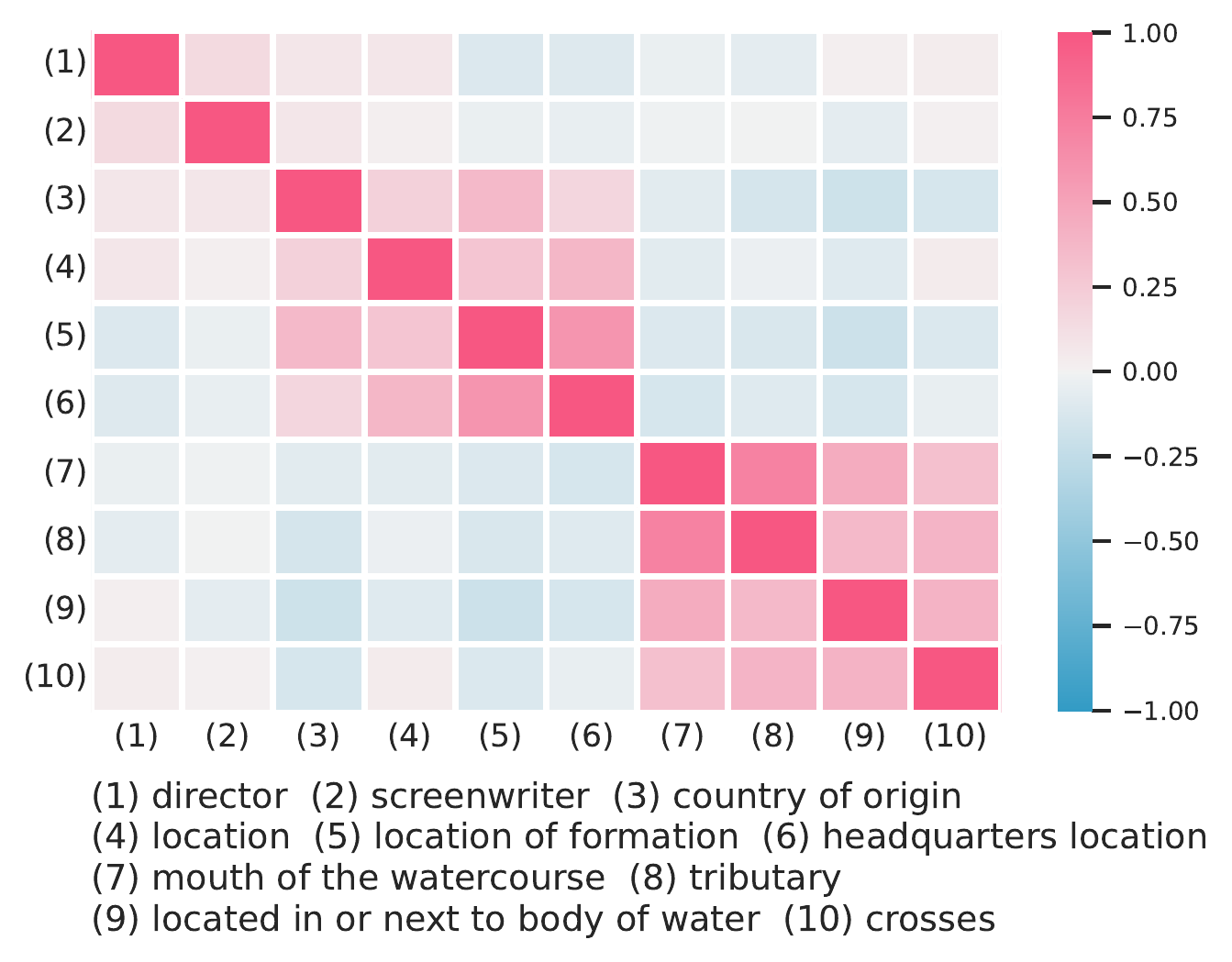}
    \caption{Visualization of cosine similarity between relation prototypes generated by our model. 
    We select 10 relations involving three highly-similar groups, i.e., [(1), (2)], [(3), (4), (5), (6)] and [(7), (8), (9), (10)].}
    \label{fig:ours-heatmap}
\end{figure}

We conduct a case study to intuitively illustrate the advantages of our model.
Figure \ref{fig:ours-heatmap} depicts the visualization result.
It is observed that the relations analogous in semantics (e.g., ``\textit{mouth of the watercourse}'' and ``\textit{tributary}'') have relatively similar relation prototypes, which reflects that our model learns a reasonable representation space.
Moreover, we see that the discrimination between similar relation prototypes (e.g., ``\textit{director}'' and ``\textit{screenwriter}'') is still obvious, which reveals that our model can distinguish analogous relations.
Please see Appendix \ref{appsec:case-study} for the comparison with CRECL.

%====================%
\section{Conclusion}
\label{sect:conclusion}

In this paper, we study continual RE. 
Through an empirical study, we find that knowledge decay among analogous relations is a key reason for catastrophic forgetting in continual RE.
Furthermore, the overfitting problem prevalent in memory-based models also lacks consideration.
To this end, we introduce a novel memory-based model to address the above issues.
Specifically, the proposed memory-insensitive relation prototypes and memory augmentation can reduce overfitting to typical samples.
In memory replay, the integrated training and focal knowledge distillation help maintain the knowledge among analogous relations, so that the model can better distinguish them.
The experimental results on the FewRel and TACRED datasets demonstrate that our model achieves state-of-the-art performance and effectively alleviates catastrophic forgetting and overfitting for continual RE.
In future work, we plan to explore whether our model can be used in few-shot RE to help distinguish analogous relations.

%====================%
\section{Limitations}

Our model may have several limitations:
(1) As a memory-based model, our model consumes additional space to store typical samples and static prototypes, which causes the performance to be influenced by the storage capacity.
(2) Although we propose memory-insensitive relation prototypes and memory augmentation, our model still relies on the selection of typical samples. 
The selected samples of low quality may harm the performance of our model.
(3) The recent progress in large language models may alleviate catastrophic forgetting and overfitting, which has not been explored in this paper yet.

%====================%
\section*{Acknowledgments}
This work was supported by the National Natural Science Foundation of China (No. 62272219) and the Collaborative Innovation Center of Novel Software Technology \& Industrialization. 

\balance
\bibliography{custom}
\bibliographystyle{acl_natbib}

%====================%
\clearpage
\appendix

\section{More Results of Empirical Study}
\label{appsec:empirical}

As mentioned in Section \ref{sect:introduction}, we conduct an empirical study to explore the causation of catastrophic forgetting and find that the knowledge among analogous relations is more likely to be forgotten.
As a supplement, we further report more results of our empirical study.
Table \ref{tab:app-empirical} shows the average change of maximum similarity when the accuracy on relations suffers a sudden drop.
Note that the number of relations greater than a 40\% drop of CRECL on the TACRED dataset is quite small, thus the result may not be representative.
It is observed that, if the maximum similarity of a relation to others obviously increases, its accuracy suddenly drops severely, which indicates that there tends to be a newly emerging relation analogous to it.
In short, we can conclude that a relation may suffer catastrophic forgetting when its analogous relations appear.
This also emphasizes the importance of maintaining knowledge among analogous relations.

\begin{table}[!h]
\resizebox{\columnwidth}{!}{
\begin{tabular}{l|c|cc}
\toprule
\multirow{2}{*}{Models} & \multirow{2}{*}{Sudden drop} & \multicolumn{2}{c}{Maximum similarity change} \\
\cmidrule(lr){3-3} \cmidrule(lr){4-4}
& & FewRel & TACRED \\
\midrule
\multirow{3}{*}{CRL} & (0.0, 20.0) & 0.715 $\rightarrow$ 0.715 & 0.780 $\rightarrow$ 0.773\\
                     & [20.0, 40.0) & 0.700 $\rightarrow$ 0.888 & 0.798 $\rightarrow$ 0.899 \\
                     & [40.0, 100.0) & 0.784 $\rightarrow$ 0.944 & 0.860 $\rightarrow$ 0.924 \\
\midrule
\multirow{3}{*}{CRECL} & (0.0, 20.0) & 0.596 $\rightarrow$ 0.601 & 0.649 $\rightarrow$ 0.642\\
                       & [20.0, 40.0) & 0.665 $\rightarrow$ 0.889 & 0.650 $\rightarrow$ 0.827 \\
                       & [40.0, 100.0) & 0.556 $\rightarrow$ 0.904 &  0.649 $\rightarrow$ 0.820 \\
\bottomrule
\end{tabular}}
\caption{More results of our empirical study. 
We report the average change of maximum similarity when the accuracy of relations suffers varying degrees of a sudden drop. 
``Sudden drop'' denotes the accuracy drop between two adjacent tasks.}
\label{tab:app-empirical}
\end{table}

\section{Implementation Details}
\label{appsec:hyperparameters}

We carry out all experiments on a single NVIDIA RTX A6000 GPU with 48GB memory. 
Our implementation is based on Python 3.9.7 and the version of PyTorch is 1.11.0.

We find the best hyperparameter values through grid search with a step of 0.1 except 0.05 for $\omega$ and 0.25 for $\gamma$.
The search spaces for various hyperparameters are $\alpha \in [0.2, 0.8], \beta \in [0.1, 0.5], \mu \in [0.1, 1.0], \omega \in [0.05, 0.25], \gamma \in [1.0, 2.0]$ and $\lambda_1,$ $\lambda_2 \in [0.5, 1.5]$.
Besides, we fix $\tau_1$ and $\tau_2$ to 0.1 and 0.5, respectively.
The used hyperparameter values are listed below: 
\begin{itemize}
\item For FewRel, $\alpha=0.5$, $\beta=0.5$, $\tau_1=0.1$, $\mu=0.5$, $\omega=0.1$, $\tau_2=0.5$, $\gamma=1.25$, $\lambda_1=0.5$, $\lambda_2=1.1$. 

\item For TACRED, $\alpha=0.6$, $\beta=0.2$, $\tau_1=0.1$, $\mu=0.8$, $\omega=0.15$, $\tau_2=0.5$, $\gamma=2.0$, $\lambda_1=0.5$, $\lambda_2=0.7$.
\end{itemize}

\section{Case Study of Our Model and CRECL}
\label{appsec:case-study}

\begin{figure}[!ht]
    \centering
    \subfigure[Visualization of our model.]{\includegraphics[width=\columnwidth]{figs/ours_heat_map.pdf}}
    \subfigure[Visualization of CRECL.]{\includegraphics[width=\columnwidth]{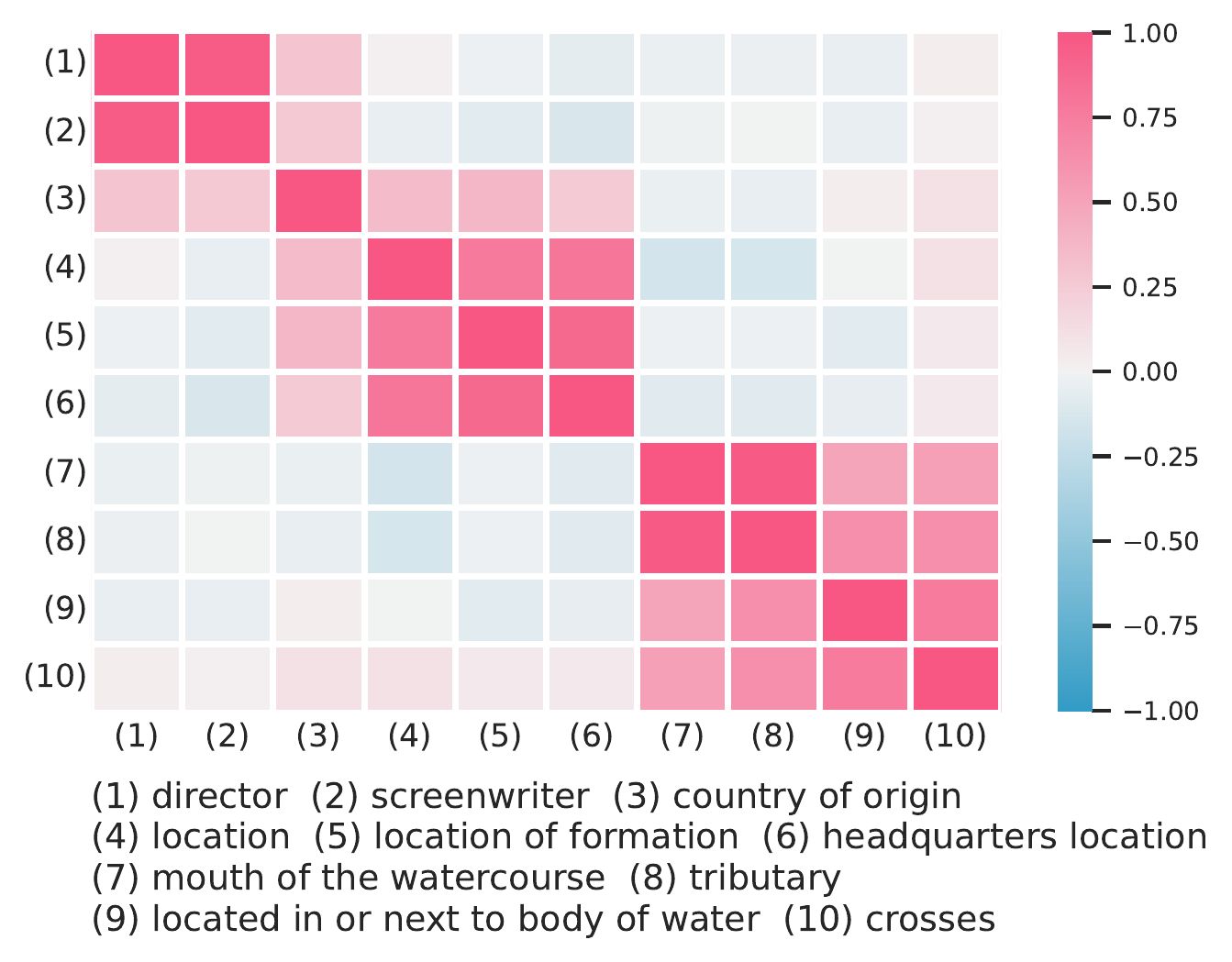}}
    \caption{Visualization of cosine similarity between relation prototypes generated by our model and CRECL.}
    \label{fig:heatmap-compare}
\end{figure}

To intuitively illustrate that our model can better distinguish analogous relations, we conduct a comparison to CRECL based on the case study in Section \ref{sec:case-study}.
As depicted in Figure \ref{fig:heatmap-compare}, it is true for both our model and CRECL that if the relations are dissimilar in semantics, the similarity between their prototypes is low.
However, we can observe that our model learns relatively dissimilar prototypes among analogous relations (e.g., lighter color between ``\textit{director}'' and ``\textit{screenwriter}''), which demonstrates that our model can better distinguish analogous relations.

\begin{table*}[!t]
\centering
{\small
\begin{tabular}{l|cccccccccc}
\toprule
  \textbf{FewRel} & $T_1$ & $T_2$ & $T_3$ & $T_4$ & $T_5$ & $T_6$ & $T_7$ & $T_8$ & $T_9$ & $T_{10}$ \\ 
\midrule
  RP-CRE + ACA & 97.7 & 95.2 & 92.8 & 91.0 & 90.1 & 88.7 & 86.9 & 86.4 & 85.3 & 83.8 \\
  EMAR + ACA & 98.3 & 94.6 & 92.6 & 90.6 & 90.4 & 88.8 & 87.7 & 86.7 & \textbf{85.6} & 84.1 \\
\midrule
  Ours & 98.1 & \textbf{95.8} & \textbf{93.6} & \textbf{91.9} & \textbf{91.1} & \textbf{89.4} & \textbf{88.1} & \textbf{86.9} & \textbf{85.6} & \textbf{84.2} \\ 
  Ours + ACA & \textbf{98.4} & 94.8 & 92.8 & 91.4 & 90.4 & 88.9 & 87.8 & 86.8 & 86.0 & 83.9 \\ 
\bottomrule
\toprule
  \textbf{TACRED} & $T_1$ & $T_2$ & $T_3$ & $T_4$ & $T_5$ & $T_6$ & $T_7$ & $T_8$ & $T_9$ & $T_{10}$ \\ 
\midrule
  RP-CRE + ACA & 97.1 & 93.5 & 89.4 & 84.5 & 83.7 & 81.0 & 79.3 & 78.0 & 77.5 & 76.5 \\
  EMAR + ACA & 97.6 & 92.4 & 90.5 & 86.7 & 84.3 & 82.2 & 80.6 & 78.6 & 78.3 & 78.4 \\
\midrule
  Ours & 97.7 & 94.3 & \textbf{92.3} & \textbf{88.4} & \textbf{86.6} & \textbf{84.5} & \textbf{82.2} & \textbf{81.1} & \textbf{80.1} & \textbf{79.1} \\ 
  Ours + ACA & \textbf{98.5} & \textbf{94.7} & 91.9 & 85.5 & 84.2 & 82.1 & 79.6 & 77.3 & 77.1 & 76.1 \\ 
\bottomrule
\end{tabular}}
\caption{Accuracy (\%) on all observed relations after learning each task.}
\label{tab:aca}
\end{table*}

\section{Comparison with ACA}
\label{appsec:aca}

As aforementioned in Section \ref{sect:literature}, \citet{ACA} propose an adversarial class augmentation (ACA) strategy, aiming to learn robust representations to overcome the influence of analogous relations.
Specifically, ACA utilizes two class augmentation methods, namely hybrid-class augmentation and reversed-class augmentation, to build hard negative classes for new tasks.
When new tasks arrive, the model is jointly trained on new relations and adversarial augmented classes to learn robust initial representations for new relations.
As a data augmentation strategy, ACA can be combined with other continual RE models.
Therefore, we conduct an experiment to explore the performance of our model with ACA.

We re-run the source code of ACA and report the results of RP-CRE + ACA, EMAR + ACA, and our model + ACA in Table \ref{tab:aca}.
Compared with the original models, both EMAR and RP-CRE gain improvement, which demonstrates the effectiveness of ACA in learning robust representations for analogous relations.
However, as we also explicitly consider the knowledge forgetting of analogous relations, there exist overlaps between ACA and our model.
Thus, the performance of our model declines when combined with ACA.
We leave the combination of our model and other augmentation methods in future work.

\section{Performance on Dissimilar Relations}

We further conduct an experiment to explore the performance on dissimilar relations.
We consider that relations with the highest similarity to other relations lower than 0.7 are dissimilar relations.
As shown in Table \ref{tab:dissim-relations}, our model achieves the best accuracy on dissimilar relations.
We attribute this to the better representations it learns through integrated training.
However, our model does not always obtain the smallest drop as it focuses on alleviating the forgetting of analogous relations.
Overall, from the results in Tables \ref{tab:ana-relations} and \ref{tab:dissim-relations}, we can conclude that our model achieves the best accuracy on both analogous and dissimilar relations as well as the least drop on analogous relations.

\begin{table}[!hb]
\centering
{\small
\begin{tabular}{l|cc|cc}
\toprule
\multirow{2}{*}{Models} & \multicolumn{2}{c|}{FewRel} & \multicolumn{2}{c}{TACRED} \\
\cmidrule(lr){2-3} \cmidrule(lr){4-5} & Accuracy & Drop & Accuracy & Drop \\
\midrule
CRL   & 90.2 & 5.9 & 92.1 & \textbf{1.4} \\
CRECL & 90.6 & 5.3 & 91.2 & 3.8 \\
Ours  & \textbf{92.4} & \textbf{4.1} & \textbf{93.7} & 2.3 \\
\bottomrule
\end{tabular}}
\caption{Accuracy (\%) and accuracy drop (\%) on dissimilar relations. 
Relations with the highest similarity to other relations lower than 0.7 are considered as dissimilar relations.
Accuracy and drop are calculated in the same way as Table \ref{tab:empirical}.}
\label{tab:dissim-relations}
\end{table}

\end{document}